\documentclass[conference]{IEEEtran}
\IEEEoverridecommandlockouts
\usepackage{cite}
\usepackage{amsmath,amssymb,amsfonts}
\usepackage{algorithmic}
\usepackage{graphicx}
\usepackage{textcomp}
\usepackage{xcolor}
\usepackage{cite}
\usepackage{balance}
\usepackage{textcomp}
\ifCLASSINFOpdf
\else
\usepackage[dvips]{graphicx}
\fi

\usepackage{algorithm}      
\usepackage{algorithmic}

\usepackage{varwidth}

\begin{document}

\title{Gaussian Process based Stochastic Model Predictive Control for Cooperative Adaptive Cruise Control}
\author{Sahand Mosharafian$^{\dag}$, Mahdi Razzaghpour$^*$, Yaser P. Fallah$^*$, Javad Mohammadpour Velni$^{\dag}$\\ \\
$^{\dag}$School of Electrical \& Computer Engineering, Univ. of Georgia, Athens, GA \\
$^*$Dept. of Electrical \& Computer Engineering, Univ. of Central Florida, Orlando, FL
}

\maketitle

\begin{abstract}
Cooperative driving relies on communication among vehicles to create situational awareness. One application of cooperative driving is Cooperative Adaptive Cruise Control (CACC) that aims at enhancing highway transportation safety and capacity. Model-based communication (MBC) is a new paradigm with a flexible content structure for broadcasting joint vehicle-driver predictive behavioral models. The vehicle's complex dynamics and diverse driving behaviors add complexity to the modeling process. Gaussian process (GP) is a fully data-driven and non-parametric Bayesian modeling approach which can be used as a modeling component of MBC. The knowledge about the uncertainty is propagated through predictions by generating local GPs for vehicles and broadcasting their hyper-parameters as a model to the neighboring vehicles. In this research study, GP is used to model each vehicle's speed trajectory, which allows vehicles to access the future behavior of their preceding vehicle during communication loss and/or low-rate communication. Besides, to overcome the safety issues in a vehicle platoon, two operating modes for each vehicle are considered; free following and emergency braking. This paper presents a discrete hybrid stochastic model predictive control, which incorporates system modes as well as uncertainties captured by GP models. The proposed control design approach finds the optimal vehicle speed trajectory with the goal of achieving a safe and efficient platoon of vehicles with small inter-vehicle gap while reducing the reliance of the vehicles on a frequent communication. Simulation studies demonstrate the efficacy of the proposed controller considering the aforementioned communication paradigm with low-rate intermittent communication.
\end{abstract}


\section{Introduction}

\noindent Adaptive cruise control (ACC) is a radar-based system, which is designed to enhance driving comfort and safety by adjusting a vehicle's speed to match the speed of the preceding vehicle. However, ACC only has a small impact on the highway capacity \cite{Effects_ACC}. The objective of cooperation in a highway scenario is to ensure that all vehicles in a lane move at the same speed while maintaining a desired formation geometry, which is specified by a desired inter-vehicle gap policy. The cooperative driving with constant spacing policy, which is called platooning, makes vehicles maintain a constant distance from their immediate predecessor while for the cooperative adaptive cruise control (CACC) constant time headway gap is used, in which the desired following distance should be proportional to the speed of the vehicle; the higher the speed, the larger the distance. CACC and platooning have the potential to increase the highway capacity when they reach a high market penetration \cite{Impacts_CACC}. It is shown that platooning is more sensitive to communication losses than the CACC is, mainly due to its very close coupling between vehicles \cite{RZ_impact}.


\begin{figure}[t]
  \centering
    \includegraphics[width=\linewidth]{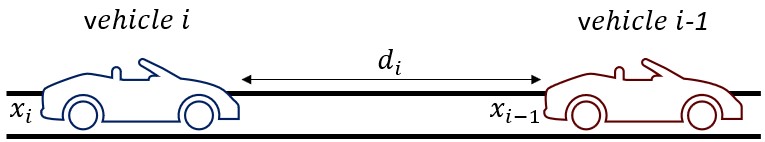}
  \caption{A simple representation of the system model. The distance between vehicle $i$ and vehicle $i-1$ is denoted by $d_i$, and $x_i$ is the location of the rear bumper of the $i^{th}$ vehicle. Vehicle $i$ receives data from its preceding vehicle through wireless communication.}
  \label{model}
\end{figure}

Model-based communication (MBC) is a recently-explored communication scalability solution, which has shown a promising potential to reduce the channel congestion \cite{model_based_communication}. The fundamental intention behind the MBC scheme is to utilize a more flexible content structure for the broadcast packets based on the joint vehicle-driver predictive behavioral models in comparison with the Basic Safety Message (BSM) content structure defined by J2735 standard. MBC can potentially shrink the payload size by extracting an abstract representation of the vehicle’s state. In addition, it reduces the transmission rate by enabling the recipient vehicles to predict their neighbors mobility more accurately for a longer future time horizon. As a result of reduced load, the MBC experiences lower rate of packet collision compared to its baseline counterpart in different traffic densities \cite{V2X_GP,Utilizing_Model_Based}.
MBC can utilize different methods of modeling vehicle movement behavior. Non-parametric Bayesian inference techniques, particularly Gaussian Processes (GPs) are among the promising methods for analytically tractable modeling of joint vehicle-driver behavior, which at the same time is not limited to some certain criteria. The driver behavioral models are functions of different factors such as the driver’s personal driving style, road traffic, weather condition, etc. Therefore, movement models may become very complex. Gaussian process regression is a powerful non-parametric tool used to infer values of an unknown function given previously collected measurements \cite{GP}. In this work, we intend to use the Gaussian process regression to derive the model of the remote vehicle and its driver as a unique object. In addition to exhibiting very good generalization properties, a major advantage of GPs is that they come equipped with a measure of model uncertainty, making them particularly beneficial for safety critical applications. The GP-based MBC module trains the GP based on the last received information. This procedure results in generating a new situational awareness messages which carry the last updated abstract model of the vehicle’s state.

The main goal of the V2V communications is to enable every vehicle in a Vehicular Ad-hoc NETwork (VANET) to frequently inform the surrounding vehicles about its most recent dynamic states. However, its performance degrades with channel load increase (i.e., high number of surrounding vehicles). Due to safety reasons, CACC methods should be robust against packet loss and communication failure \cite{ploeg2014analysis}. CACC would degrade to ACC if no action is taken to compensate for packet losses \cite{ploeg2014graceful}. According to \cite{wu2019cooperative}, three general solutions are available to cope with the safety problem during communication failure:
\begin{itemize}
\item Improving the communication protocol (e.g., see \cite{lee2002packet}); \item Increasing the vehicle headway during failure (e.g., see \cite{abou2017adaptive}); 
\item Adapting CACC algorithms to cope with the safety problem (e.g., using a model predictive control approach).
\end{itemize}
The method proposed in this work aims at improving the robustness of the controller to the communication losses, and successfully operating in low-rate communication scenarios. The method proposed in this paper fuses the first and third solutions by taking advantage of two important factors, namely, model-based communication and emergency braking.

In this paper, the CACC design problem consists of two operating modes for each vehicle: free following and emergency braking. In free following mode, each vehicle simply follows its preceding vehicle while the vehicle in the emergency braking mode performs hard braking to avoid possible collision and preserve safety. These operating modes add binary variables to the system, and MBC contains stochastic information about preceding vehicle speed profile. To formulate the underlying problem, a hybrid stochastic model is needed. Hence, discrete hybrid stochastic automata, introduced initially in \cite{bemporad2010model} suits the aforementioned stochastic system with both continuous and discrete variables. Finally, each vehicle employs Model Predictive Control (MPC) to optimize their speed trajectory considering system constraints. By employing MBC, during communication loss and low-rate communication, CACC will not degrade to ACC since the preceding vehicle trajectory can be produced through GP-based MBC, which assures the efficacy and the safety of the system even with a low communication rate. 

The contributions of the paper are as follows. The paper fuses model-free and model-based communication paradigms to assure the CACC safety and performance during communication losses. At each successful communication event, vehicles directly share their future speed profile, and they also share/update a model which allows vehicles to predict their preceding vehicle's behavior so that each autonomous vehicle is able to plan its speed profile properly. Besides, integrating two operating modes (namely free following and emergency braking) improves safety while avoiding unnecessary braking through penalizing excessive braking.

The rest of the paper is organized as follows. The system model and Gaussian process are described in Section II. System model considering constraints and operating modes are presented in mixed logical dynamical from in Section III. The underlying MPC problem is then explained in Section IV. The performance of the proposed controller is evaluated through simulation studies in section V, and concluding remarks are finally made in Section VI.

\section{Preliminaries and System Model}

\noindent This section first provides a state-space representation of the vehicle model used for the purpose of CACC design. Then, an introduction to Gaussian processes is given.

\subsection{System Model}

\noindent In this paper, we aim at designing controllers for CACC that build a string stable vehicle platoon with longitudinal movement while reducing the reliance of the system on frequent communication. For safety reasons, emergency braking is taken into account to avoid possible collisions that might occur due to the sudden changes in the preceding vehicle's speed or random packet loss. The CACC enables reducing the distance between CAVs in a platoon thereby increasing the road capacity while preserving safety.

A CACC system with $N_v$ vehicles is considered here, and the index $i\in\{0,1,\hdots,N_v-1\}$ is used to represent the $i^{th}$ vehicle with $i=0$ being the leader vehicle. The distance between $i^{th}$ vehicle and its preceding vehicle at time $t$ is denoted by $d_i(t)$ (see Fig. \ref{model}) and defined as
\begin{equation} \label{d}
    \begin{aligned}
        d_{i}(t) = x_{i-1}(t)-x_{i}(t)-l_{i},
    \end{aligned}
\end{equation}
where $x_i$ and $l_{i}$ are the location of the $i^{th}$ vehicle rear bumper, and the length of the $i^{th}$ vehicle, respectively. Hence, the desired spacing policy for vehicle $i$ shown by $d^{*}_{i}(t)$ can be defined as
\begin{equation} \label{d*}
    \begin{aligned}
        d^{*}_{i}(t) = \tau\,v_{i}(t)+d^{s}_{i},
    \end{aligned}
\end{equation}
where $v_i(t)$ is the vehicle speed at time $t$, $\tau$ is the time gap, and $d^{s}_{i}$ is the stand still distance. Using constant time headway gap as in \eqref{d*} improves string stability and safety\cite{naus2010string}. The difference between the distance and its desired value is defined as
\begin{equation} \label{dd}
    \begin{aligned}
        \Delta d_{i}(t) = d_{i}(t)-d^{*}_{i}(t),
    \end{aligned}
\end{equation}
and speed difference between the $i^{th}$ vehicle and its preceding vehicle is considered as $\Delta v_{i}(t) = v_{i-1}(t)-v_{i}(t)$ and therefore $\Delta\dot{d}_i(t) = \Delta v_{i}(t)-\tau\,a_i(t)$ and $\Delta\dot{v}_{i}(t) = a_{i-1}(t)-a_i(t)$, where $a_i(t)$ is the acceleration of  the $i^{th}$ vehicle at time $t$. Let us consider a linear model for the $i^{th}$ vehicle as
\begin{equation} \label{adot}
    \begin{aligned}
        \dot{a_{i}}(t) = -\mathnormal{f}_{i} a_{i}(t) + \mathnormal{f}_{i}u_{i}(t),
    \end{aligned}
\end{equation}
where the constant $\mathnormal{f}_i$ represents drive-line dynamics and its unit is $s^{-1}$. State-space representation of the given system for the $i^{th}$ vehicle is as follows
\begin{multline}  \label{css}
        \dot{\mathbf{x}}_i(t)=A_i\,\mathbf{x}_i(t)+B_i\,u_i(t)+D\,a_{i-1}(t)\\ \\=
        \begin{bmatrix}
            0&1&-\tau \\0&0&-1\\0&0&-\mathnormal{f}_{i}
        \end{bmatrix}\mathbf{x}_i(t)+
        \begin{bmatrix}
            0\\0\\\mathnormal{f}_{i}
        \end{bmatrix}u_{i}(t)+
        \begin{bmatrix}
            0\\1\\0
        \end{bmatrix}a_{i-1}(t)
\end{multline}
where $\mathbf{x}_i(t)=[\Delta d_i(t)\,\,\Delta v_{i}(t) \,\, a_i(t)]^{T}$. For the leader ($i=0$), $a_{i-1}(t)$ in \eqref{css} can be assumed to be zero.

Using forward-time approximation for the first-order derivative, \eqref{css} can be written in discrete-time form. The discrete-time state space model for each follower is as follows
\begin{multline}\label{fdss}
        \mathbf{x}_i(k+1) = (I+t_s\,A_i)\,\mathbf{x}_i(k)+t_s\,B_i\,u_i(k)+t_s\,D\,a_{i-1}(k)
\end{multline}
where $I$ is the identity matrix and $t_s$ is the sampling time.

\subsection{Gaussian Process}

\noindent A Gaussian process is a collection of random variables, any finite number of which have a joint Gaussian distribution. Gaussian process regression (GPR) is a non-parametric Bayesian approach to provide uncertainty measurements on the predictions. A Gaussian process is completely specified by its mean function and covariance function, which is called the kernel function in the GP context and defines the trend of target function based on the similarity pattern among the observed values. We define mean function $m(t)$ and the kernel function $k(t, t^{\prime})$ of a real process $\mathcal{V}(t)$ as
\begin{equation}
m(\mathbf{t})=\mathbb{E}[\mathcal{V}(\mathbf{t})]
\label{mean_gp}
\end{equation}
\begin{equation} \label{kernel}
k\left(\mathbf{t}, \mathbf{t}^{\prime}\right)=\mathbb{E}\left[(\mathcal{V}(\mathbf{t})-m(\mathbf{t}))\left(\mathcal{V}\left(\mathbf{t}^{\prime}\right)-m\left(\mathbf{t}^{\prime}\right)\right)\right]
\end{equation}
and write the Gaussian process as
\begin{equation}
\mathcal{V}(\mathbf{t}) \sim \mathcal{G} \mathcal{P}\left(m(\mathbf{t}), k\left(\mathbf{t}, \mathbf{t}^{\prime}\right)\right).
\end{equation}

Different vehicle dynamics can be considered as separate time series which should be regressed using an appropriate supervised learning method. The regression problem here is equivalent to inferring the characteristics of the unknown target functions which have generated these time series. In our case, the random variables represent the speed of the vehicle at time $t$. We are interested in incorporating the knowledge that the training data provides about the function and its future values. The joint distribution of the past values, $\mathcal{V}$, and the future values $\mathcal{V}^{\ast}$ according to the prior is
\begin{equation}
\left[\begin{array}{l}
\mathbf{\mathcal{V}} \\
\mathbf{\mathcal{V}}^{*}
\end{array}\right] \sim \mathcal{N}\left(\mathbf{0},\left[\begin{array}{ll}
K(t, t) & K\left(t, t^{*}\right) \\
K\left(t^{*}, t\right) & K\left(t^{*}, t^{*}\right)
\end{array}\right]\right).
\end{equation}
If there are $n$ training points and $n^{\ast}$ test points, then $K(X, X^{\ast})$ denotes the $n \times n^{\ast}$ covariance matrix evaluated at all pairs of training and test points, and similarly for the other entries $K(X, X)$, $K(X^{\ast},X^{\ast})$ and $K(X^{\ast}, X)$. To get predictions at unseen time of interest $t^{\ast}$, the predictive distribution can be calculated by weighting all possible predictions by their calculated posterior distribution or in probabilistic terms
\begin{equation}
\begin{aligned}
\mathbf{\mathcal{V}}^{*} \mid t^{*}, t, \mathbf{\mathcal{V}} \sim \mathcal{N} &\left(K\left(t^{*}, t\right) K(t, t)^{-1} \mathbf{\mathcal{V}}, \right.\\ K\left(t^{*}, t^{*}\right) &\left. -K\left(t^{*}, t\right) K(t, t)^{-1} K\left(t, t^{*}\right)\right).
\end{aligned}
\label{eq:probabilistic}
\end{equation}
Function values $\mathcal{V}^{\ast}$ (corresponding to test inputs $t^{\ast}$) can be sampled from the joint posterior distribution by evaluating the mean and covariance matrix from \eqref{eq:probabilistic}.

We choose the most commonly used kernel in machine learning which is the Gaussian form radial basis function (RBF) kernel. It is also commonly referred to as the exponentiated quadratic or squared exponential kernel. The kernel’s parameters are estimated using the maximum likelihood principle. The advantage of GP is its capability to substantially increase the forecasting accuracy over longer prediction horizons without reception of new raw information or model updates during a certain period. The computational complexity of a GP regression strongly depends on the number of data points $N$. Training window size has been set to 5 latest equally spaced (last 0.5 second) observed/received speed samples in time.

\section{Vehicle Model and Constraints in Mixed Logical Dynamical Form}

\noindent According to \cite{bemporad2010model}, a stochastic system with both binary and continuous/discrete-time variables and inputs can be modeled using a discrete hybrid stochastic automata (DHSA) which consists of four components: a switched affine model, an event generator, a mode selector, and a finite state machine. Representing a system in DHSA is explained in details in \cite{bemporad2010model}. CACC problem in this paper consists of two operating modes: free following and emergency braking. The system is modeled using DHSA in which the uncontrollable events are defined in terms of uncertainty in speed prediction using GP regression. Then, DHSA is presented in mixed logical dynamical (MLD) form \cite{bemporad1999control}, and finally, mixed-integer quadratic programming is employed to find optimal control input(s). In the remainder of this section, system inequalities in the form of an MLD are derived.

The constraints on the system include bounds on the acceleration, input, road speed limit, and distance between vehicles (note that a negative distance implies collision and therefore should not occur). The following inequalities should always hold true
\begin{subequations} \label{bounds}
    \begin{gather}
        \label{accbound}
        a_i^{min}\leq a_i(k)\leq a_i^{max},\\ 
        \label{inputbound}
        u_i^{min}\leq u_i(k)\leq u_i^{max},\\
        \label{speedlim}
        v_i(k)\leq v_{max},\\
        d_i(k)>0.
    \end{gather}
\end{subequations}
Besides, for passenger comfort, system input changes are bounded as follows
\begin{equation}
    \begin{gathered}
        t_s\,u_i^{min} \leq u(k+1)-u(k) \leq t_s\,u_i^{max}.
    \end{gathered}
    \label{passenger_comfort}
\end{equation}

To determine the mode of the system, the auxiliary binary variable $\xi_{i}^{e}(k)$ is considered such that
\begin{equation*}
    \begin{aligned}
        \xi_{i}^{e}(k)=1 \iff \text{Emergency braking constraint is activated}.
    \end{aligned}
\end{equation*}
First, emergency braking constraint needs to be defined. Here, for safety purpose and collision avoidance, a constraint is considered such that if $\Delta d_i(k)$ goes below a fixed level ($\,\underline{d}\,$), $i^{th}$ vehicle should brake with the minimum possible acceleration ($u_i(k)=u_i^{min}$), thereby
\begin{equation}
    \begin{aligned} \label{ems}
        \Delta d_i(k)+\underline{d} \leq 0 \iff \xi_{i}^{e}(k)=1,
    \end{aligned}
\end{equation}
which can be described in MLD form using the following inequalities
\begin{equation}
    \begin{gathered}
          \Delta d_i(k)+\underline{d}\leq \overline{b_i^{e}}[1-\xi_i^e(k)],\\
          \Delta d_i(k)+\underline{d}\geq \varepsilon+\xi_i^e(k)[\underline{b_i^e}-\varepsilon],
    \end{gathered}
\end{equation}
where $\varepsilon$ is the machine precision, and
\begin{equation*}
    \begin{gathered}
          \underline{b_i^e}\leq \Delta d_i(k)+\underline{d}\leq \overline{b_i^{e}}.
    \end{gathered}
\end{equation*}

Next, to enforce hard braking, an upper bound constraint on $u_i(k)$ is added such that
\begin{equation}
    \begin{aligned} \label{u_l1}
        u_i(k)\leq\xi_{i}^{e}(k)\,u_i^{min} +[1-\xi_{i}^{e}(k)]\,u_i^{max}.
    \end{aligned}
\end{equation}
Thus, as far as the system operates in the free following mode, the upper bound on input is $u_i^{max}$. However, when emergency braking mode is activated, the upper bound becomes $u_i^{min}$. Since the emergency braking mode forces the input $u_i(k) = u_i^{min}$, the speed of the vehicle may become negative. To handle this issue, the constraint $v_i(k) \geq 0$ is added to the system. However, this may result in infeasibility because of relation of speed at time $k$ and $k+1$; $v_i(k)\geq0$ while $v_i(k+1)=v_i(k)+t_s\,a_i(k)$ may become negative. To handle this edge case, two new auxiliary binary variables $\xi_{i}^{v}(k)$ and $\xi_{i}^{E}(k)$ are defined, and \eqref{u_l1} rewritten as
\begin{equation}
    \begin{gathered}  \label{u_l2}
        u_i(k)\leq\,[\xi_{i}^{E}(k)]\,u_i^{min} +[1-\xi_{i}^{E}(k)]\,u_i^{max},
    \end{gathered}
\end{equation}
where 
\begin{subequations} \label{EMS_FINAL}
    \begin{gather}
    \label{f_1}
    v(k)<\underline{v}_i\iff \xi_{i}^{v}(k)=0,\\
    \label{f_2}
    \xi_{i}^{E}(k)=\xi_{i}^{e}(k)\,\xi_{i}^{v}(k).
    \end{gather}
\end{subequations}
This implies that if the emergency braking event is activated but the vehicle speed is below the threshold, there is no need to enforce the vehicle input to get to its minimum. With this assumption and by simply choosing $\underline{v}_i=1\,m/s$, this issue is addressed. Consequently, the vehicle performs emergency braking if and only if $\xi_{i}^{e}(k)\,\xi_{i}^{v}(k)=1$. The statement in \eqref{f_1} in the MLD form turns into
\begin{equation*}
    \begin{gathered}
        v_i(k)-\underline{v}_i \leq \overline{b_i^v}\,\xi_{i}^{v}(k),\\
        v_i(k)-\underline{v}_i \geq \varepsilon +  [1-\xi_{i}^{v}(k)](\underline{b_i^v}-\varepsilon),
    \end{gathered}
\end{equation*}
where
\begin{equation*}
    \begin{gathered}
        \underline{b_i^v}\leq v_i(k)-\underline{v}_i \leq \overline{b_i^v},
    \end{gathered}
\end{equation*}
while \eqref{f_2} is equivalent to the following inequalities
\begin{equation}
    \begin{gathered}
        \xi_{i}^{E}(k)\geq \xi_{i}^{e}(k)+\xi_{i}^{v}(k)-1,\\
        \xi_{i}^{E}(k)\leq \xi_{i}^{e}(k),\\
        \xi_{i}^{E}(k)\leq \xi_{i}^{v}(k).
    \end{gathered}
\end{equation}
Whenever a vehicle performs emergency braking, the constraints for passenger comfort should be ignored. Hence, \eqref{passenger_comfort} is updated as follows
\begin{equation}
    \begin{gathered}
         (1-\xi_{i}^{E}(k))\,t_s\,u_i^{min}-\xi_{i}^{E}(k)\,\bar{u_i} \leq u(k+1)-u(k),\\ 
        u(k+1)-u(k) \leq \xi_{i}^{E}(k)\,\bar{u_i}+ (1-\xi_{i}^{E}(k))\,t_s\,u_i^{max},
    \end{gathered}
\end{equation}
where $\bar{u_i}=u_i^{max}-u_i^{min}$.

As described earlier, each vehicle needs to access the preceding vehicle's future speed trajectory (or the future acceleration profile), which is available in either of two ways; each vehicle will receive it through communication, or the vehicle will generate the profile using the last GP model received from the preceding vehicle. In the case that the speed profile is generated using GP, the variance of the GP output should be also taken into account. Hence, the uncertain part of the future velocity (the kernel introduced in \eqref{kernel}) can be discretized to some levels  $\{n_{i-1}^1(k),n_{i-1}^2(k),...,n_{i-1}^{m_{i-1}}(k)\}$ with known probabilities $\{p_{i-1}^1,p_{i-1}^2,...,p_{i-1}^{m_{i-1}}\}$. Therefore, the preceding vehicle velocity uncertainty can be formulated as
\begin{equation*}
    \begin{gathered}
        \eta_{i-1}(k)=
        \begin{bmatrix}
            n_{i-1}^1 & n_{i-1}^2 & \hdots & n_{i-1}^{m_{i-1}}
        \end{bmatrix}
        \begin{bmatrix}
            w_{i-1}^1(k)\\w_{i-1}^2(k)\\ \vdots \\ w_{i-1}^{m_{i-1}}(k)
        \end{bmatrix},
    \end{gathered}
\end{equation*}
where $w_{i-1}^j(k)\,,j\in\{1,2,\hdots,m_{i-1}\}$ are auxiliary binary variables that represent uncontrollable events and $P[w_{i-1}^j(k)=1]=p_{i-1}^j$. Based on the above parametrization of the uncertainty, the following equality should always hold true
$$
        \sum_{j=1}^{m_{i-1}}w_{i-1}^j(k)=1.
$$

Using the auxiliary variables added to the system, the discrete-time state space model in \eqref{fdss} is reformulated as
\begin{equation} \label{finaleq}
\begin{gathered}
    \mathbf{x}_i(k+1) =
    \begin{bmatrix}
            1 & t_s &-\tau\,t_s\\
            0 & 1 & -t_s\\
             0 & 0 & 1-t_s\mathnormal{f}_{i}
        \end{bmatrix}\mathbf{x}_i(k)+
        \begin{bmatrix}
            0\\0\\t_s\mathnormal{f}_i
        \end{bmatrix}{u}_{i}(k)\\~~~~~+
        \begin{bmatrix}
            0\\1\\0
        \end{bmatrix}\eta_{i-1}(k)+
        \begin{bmatrix}
            0\\t_s\\0
        \end{bmatrix}{a}_{i-1}(k).
    \end{gathered}
\end{equation}
If the acceleration profile is received through communication, constants $n_{i-1}^j(k)$ will be simply set to zero (the problem turns into discrete hybrid automata). Whenever GP model is used (between communication events or during communication loss), $n_{i-1}^j(k)$ values are chosen based on the GP output, and $a_{i-1}(k)$ can be easily calculated as $\left[\hat{v}_{i-1}(k+1)-\hat{v}_{i-1}(k)\right]/t_s$ where $\hat{v}_{i-1}(k)$ is the mean of the GP in \eqref{mean_gp}.

\section{Discrete Hybrid Stochastic MPC Problem for CACC Design}

\noindent To achieve a string stable CACC system, $\mathbf{x}_i(k)$ should converge to zero for every follower vehicle. This implies that the distance between $i^{th}$ vehicle and its preceding vehicle converges to its desired value while vehicles move with equal and constant speed. For the CACC design problem, when the number of vehicles increases, centralized MPC is not computationally efficient\cite{zhou2019distributed}. Besides, communication losses would make centralized MPC unreliable and even dangerous for such safety-critical systems. Instead, distributed MPC can be used to reach a string stable CACC. The MPC problem for each vehicle is formulated as follows
\begin{multline} \label{MPn_prob}
        \min_{\textbf{u}_i,\textbf{w}_{i-1},\textbf{z}_i}\\ \sum_{k=0}^{N-1}\Big[(\mathbf{x}_i(k)-R_i)^T\, Q_i \,(\mathbf{x}_i(k)-R_i)\Big]-q_i\, \ln(\pi(\textbf{w}_{i-1}))\\
        \text{subject to:}~~~~~
        \text{MLD system equations,}\\ \ln(\pi(\textbf{w}_{i-1}))\geq \ln(\tilde{p_i})~~~~~~~~
\end{multline}
where $\textbf{w}_{i-1}$ is a vector including all uncontrollable events, $\textbf{u}_i$ and $\textbf{z}_i$ are the system inputs and the vector of auxiliary variables from $k=0$ to $k=N-1$, respectively, and $\pi(\mathbf{w}_{i-1})$ denotes the probability of a trajectory occurred by $\mathbf{w}_{i-1}$ (see \cite{bemporad2010model} for details).

In this paper, the CACC leverages the model-based communication in which vehicles share their latest GP model with their follower vehicle. It is assumed that each vehicle uses a low-rate communication to share its future acceleration profile (model-free communication) with its follower vehicle. Consequently, every vehicle has access to its preceding vehicle's future speed trajectory through either the communication (every $t_c$ seconds if packet loss does not occur) or the GP model (until the next successful communication event). Consequently, during communication loss or low-rate data exchange, CACC will not degrade to ACC. If a vehicle uses the GP model, its optimization problem is a discrete hybrid stochastic MPC while if the future speed profile is received through communication, the optimization problem turns into a discrete hybrid MPC. For simplicity, the controller that solely relies on the model-free communication is called DHMPC, the controller that only relies on the model-based communication is called DHSMPC, and the one that takes advantage of both types of communication is named DH-DHSMPC.

\section{Simulations Results and Discussion}

\noindent Simulations are conducted considering 10 vehicles and using multiple scenarios. CVXPY package in Python is used for implementing the optimization problem and Gurobi optimization package is used as the solver for the mixed integer programs \cite{diamond2016cvxpy, agrawal2018rewriting,gurobi}. The desired speed trajectory for the leading vehicle in the simulation is considered to be
\begin{equation}
    \begin{gathered}
    v^*_0(t)=
    \begin{cases}
    27 &\,\,\,\, t<15\,s,\\
    0 &\,\,\,\, 15\,s\leq t < 30\,s,\\
    25 &\,\,\,\, t \geq 30\,s.
    \end{cases}
    \end{gathered}
\end{equation}
Parameters used in these simulations can be found in Table \ref{table1}. Three different case studies are investigated here. In the first two case studies, the performance of the pure DHSMPC (using GP model, solving discrete hybrid stochastic MPC) is compared with the DHMPC (sharing future acceleration profile through communication, solving a discrete hybrid MPC). Finally, the performance of the DH-DHSMPC, which leverages both model-based and model-free communication, is compared with DHMPC when communication rate is low (1 Hz) and an \textit{iid} packet loss is considered.

\begin{table}[h]
\caption{Model and optimization parameters used in the simulations.}
\centering
\renewcommand{\arraystretch}{1.5}
\begin{tabular}{||p{1cm}|p{2.45cm}||p{1cm}|p{2.45cm}||} 
 \hline\hline
 parameter & value & parameter & value\\ [0.5ex] 
 \hline\
 $N$ &  $7$ & $t_s$ & $0.1\,s$
 \\\
 $l_i$ & $5\,m$ & $d^s_i$ & $2\,m$ \\\
 $\underline{d}$ & $1\, m$ & $f_i$ & $10\,s^{-1}$\\\
 $a_i^{min}$ & $ -4\,m/s^2$ &  $a_i^{max}$ & $ 3\,m/s^2$\\ 
  $u_i^{min}$ & $ -4\,m/s^2$ &  $u_i^{max}$ & $ 3\,m/s^2$\\ 
 $\hat{p}$ & $0.01^{N}$ & $q_i$ & $10$\\[1ex] 
 \hline\hline
\end{tabular}
\label{table1}
\end{table}

\begin{figure}[t]
  \centering
    \includegraphics[width=\linewidth]{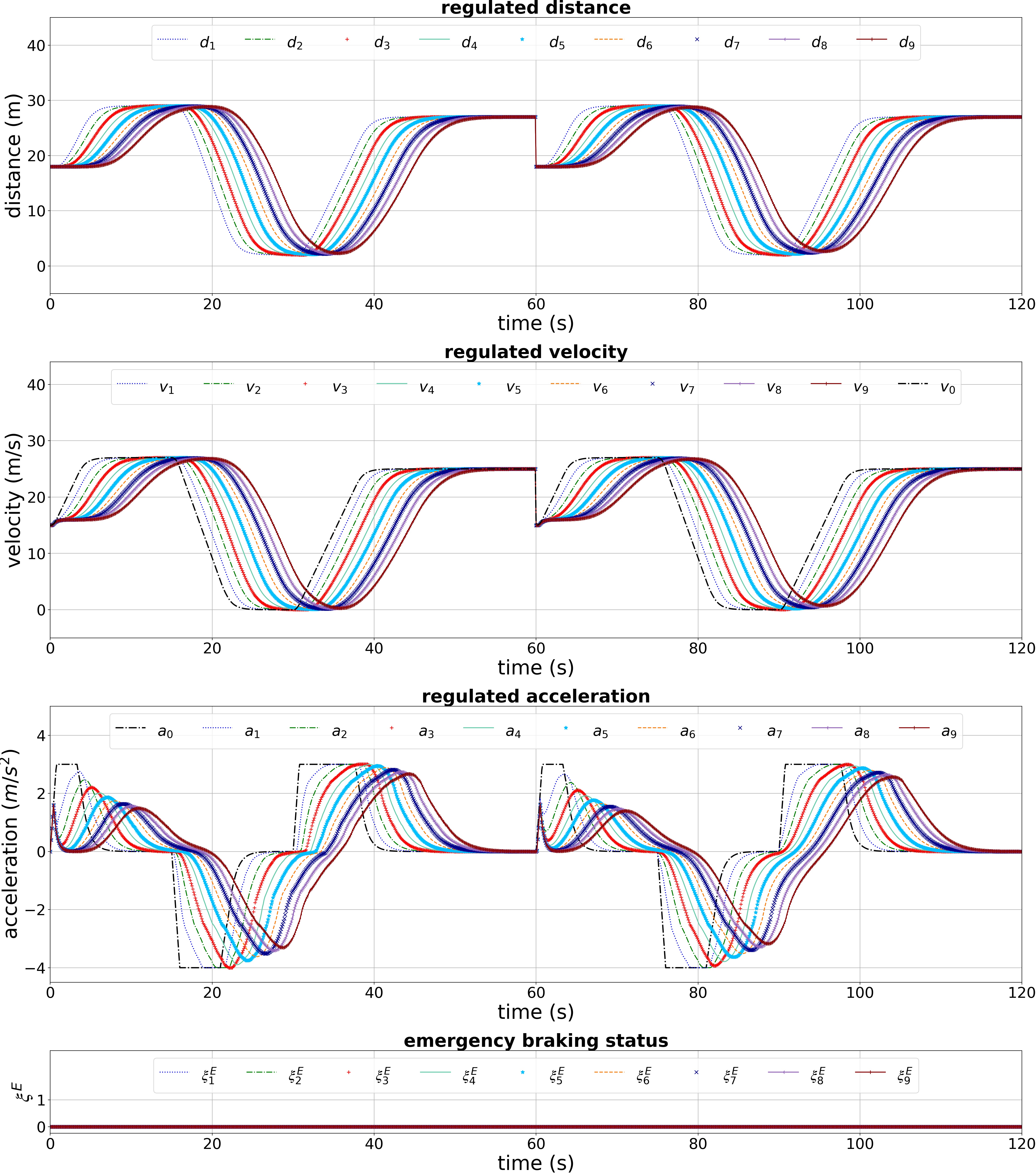}
  \caption{Comparing the performance of the DHSMPC (GP-based communication) with DHMPC (model-free communication) when the time gap of $1\,s$ is considered. From time $0\,s$ to $60\,s$, the platoon operates based on the GP model while the environment resets at time $60\,s$ and after that vehicles share their speed trajectory. It is observed that the performance of both methods is almost the same, which demonstrates the efficacy of the GP-based method.}
  \label{DHSA_Comparision_T_1}
\end{figure}

\begin{figure}[t]
  \centering
    \includegraphics[width=\linewidth]{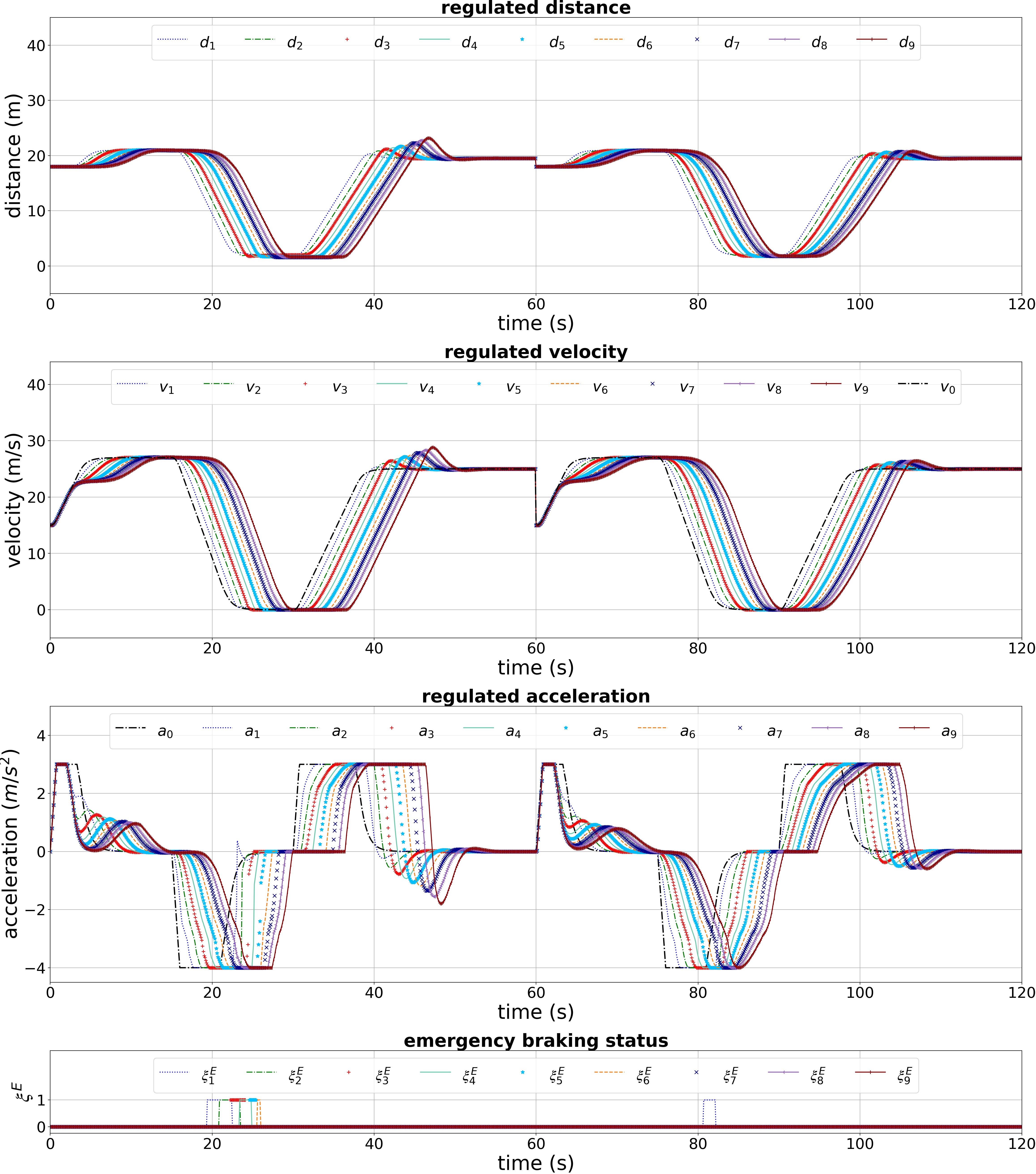}
  \caption{Comparing the performance of the DHSMPC (GP-based communication) with DHMPC (model-free communication) when the time gap is $0.7\,s$. From time $0\,s$ to $60\,s$, the platoon operates using the GP model while the environment resets at time $60\,s$ and after that vehicles share their speed trajectory. Although there are overshoots during acceleration in the GP-based method and emergency braking occurs more frequently during deceleration, the vehicles are still able to operate safely.}
  \label{DHSA_Comparision_T_0.7}
\end{figure}

\begin{figure}
  \centering
    \includegraphics[width=\linewidth]{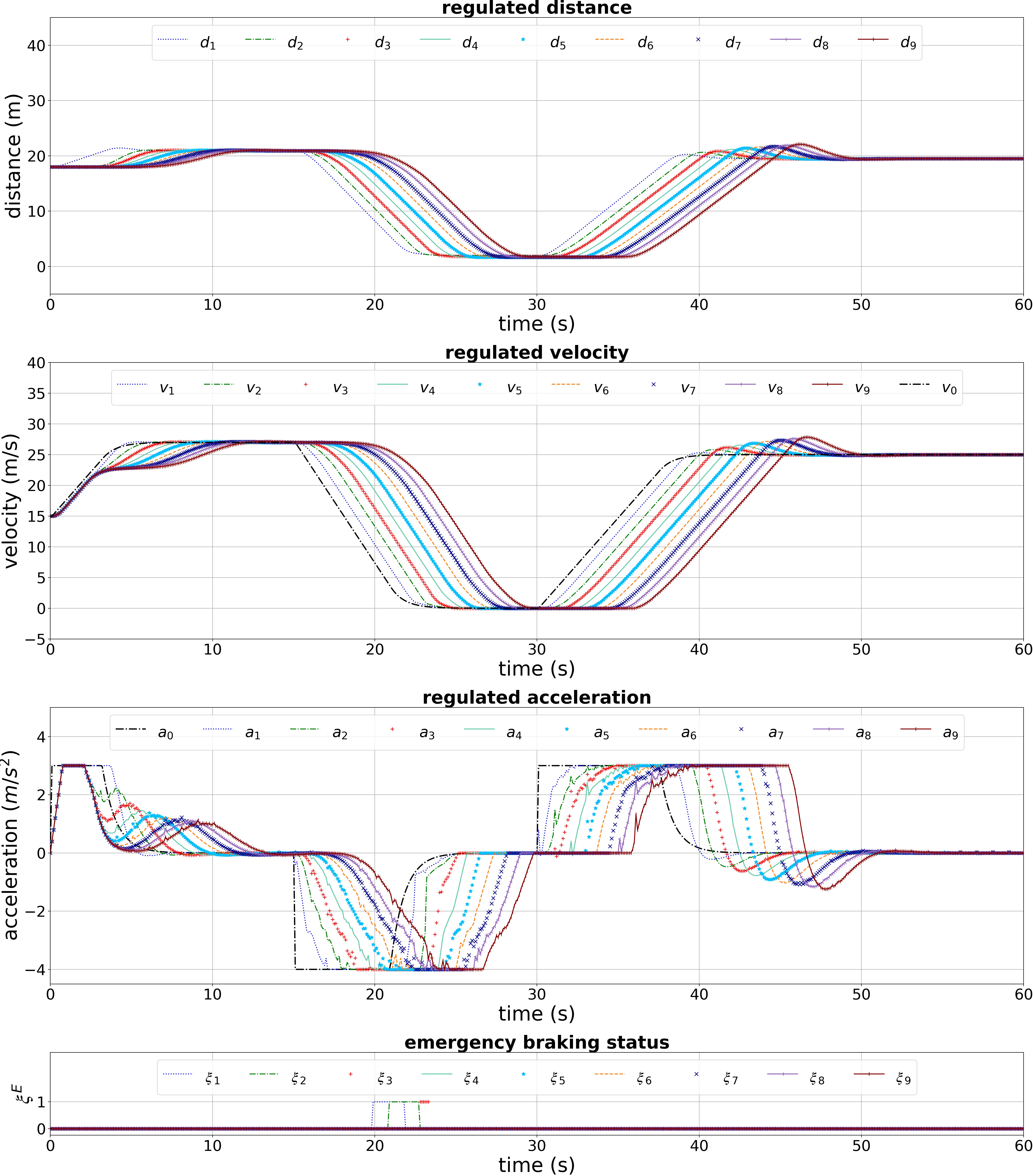}
  \caption{Performance of the DH-DHSMPC with 10 vehicles, $T=0.7\,s$, $t_c=1\,s$, and communication loss with the probability of 0.25. Using the GP model results in compensating both the low-rate communication and communication failure.}
  \label{DHSA_GP_T_0.7_loss_0.75}
  \vspace{0.67cm}
    \includegraphics[width=\linewidth]{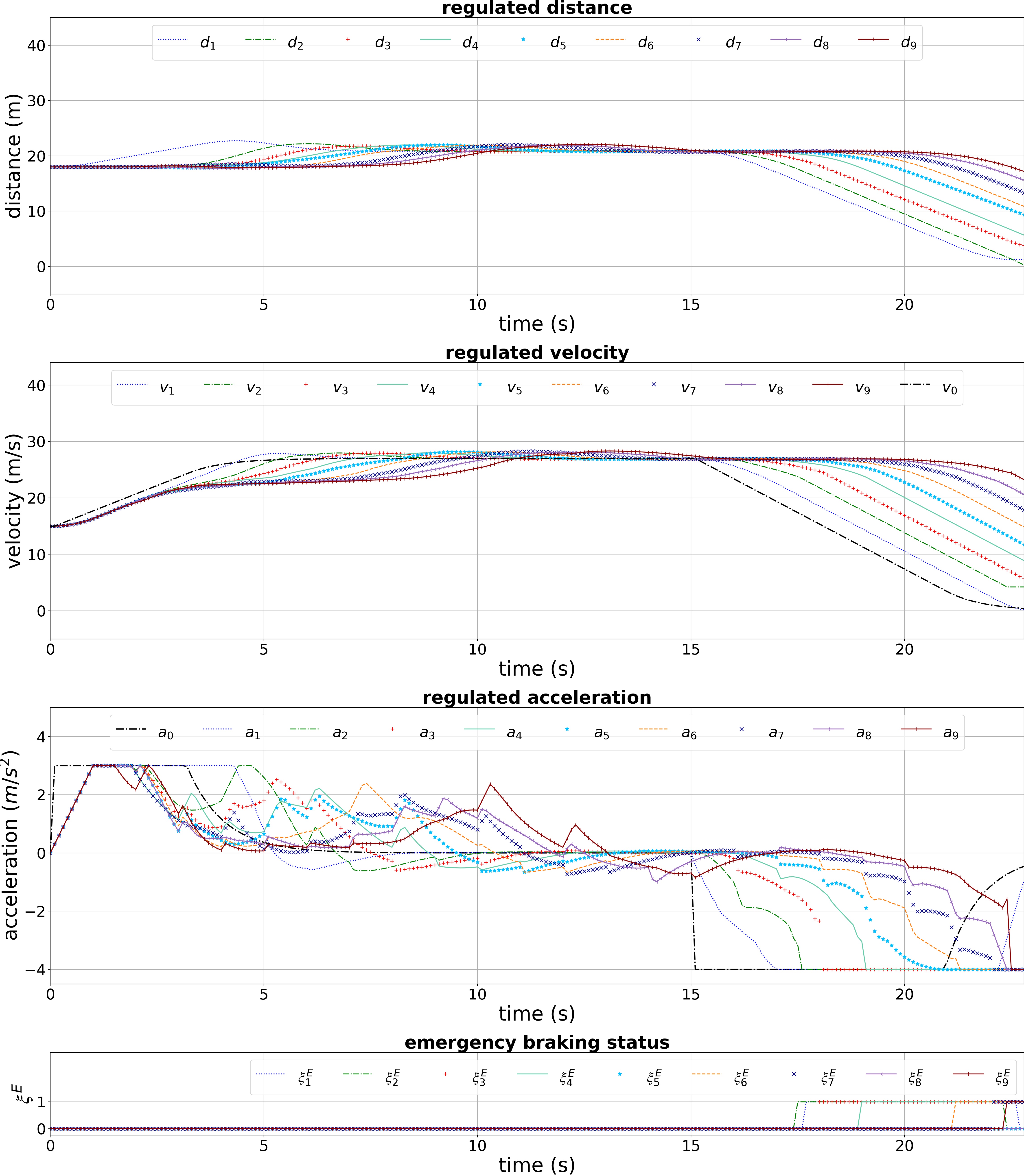}
  \caption{Performance of the DHMPC with 10 vehicles, $T=0.7\,s$, $t_c=1\,s$, and communication loss with the probability of 0.25. As observed, the vehicles are not able to cope with the sudden deceleration, and an accident occurs.}
  \label{DHSA_accident}
\end{figure}

For the DHMPC case, each vehicle employs the time-shifted acceleration information of its preceding vehicle to efficiently use the last received data between two successful communication events. Assuming that the last communication occurred at time instant $k_0$, at $k_1<k_0+N-1$, ${a}_{i-1}(k)$ in \eqref{finaleq} is replaced by
\begin{multline}
    \hat{a}_{i-1}(k)=
    \begin{cases}
    a_{i-1}(k) & \,k_1\leq k \leq k_0+N-1\\ 
    b(k) & \, \text{otherwise} \\
    \end{cases}
\end{multline}
where $b(k)$ can be estimated using a linear extrapolation of the last two samples of $a_{i-1}(k)$. If there is a communication loss and $i^{th}$ vehicle does not receive information from its preceding vehicle, then the vehicle would resort to operating in adaptive cruise control (ACC) mode until it receives information from its preceding vehicle and returns back to CACC mode.

In Figs. \ref{DHSA_Comparision_T_1}-\ref{DHSA_accident}, the first subplot shows every vehicle's distance from its predecessor ($d_i(t)$) while the second subplot shows each vehicle's velocity ($v_i(t)$). The third subplot depicts each vehicle's acceleration information, and the last one shows the emergency braking status ($\xi^E_i(t)$). As shown in Fig. \ref{DHSA_Comparision_T_1}, in the first case study, when vehicles try to keep a time gap of $1\,s,$ the performance of the both DHSMPC and DHMPC is almost the same, and the vehicles do not need to perform emergency braking, which shows the efficacy of the model-based communication. To evaluate the performance of the model-based communication (DHSMPC) with smaller vehicle headway, the time gap is reduced to $0.7 \,s$ in the second case study. As shown in Fig. \ref{DHSA_Comparision_T_0.7}, the vehicles with the proposed communication paradigm are able to safely follow their preceding vehicle. However, compared to the model-free communication case, vehicles need to perform emergency braking frequently during deceleration, and there are noticeable overshoots during acceleration in the distance/velocity profiles.

In the last simulation study, the performance of DHMPC is compared with DH-DHSMPC. In this scenario, the time gap is $0.7\,s$, and vehicles send their predictive acceleration trajectory every $1\,s$ ($t_c=1$) while the communication success probability is $0.75$. Fig. \ref{DHSA_GP_T_0.7_loss_0.75} shows the performance of the DH-DHSMPC method. Since this method takes advantage of the model-based communication, vehicles are able to operate safely and efficiently with a low communication rate and even in the presence of a communication failure. During the deceleration event, a few vehicles enter the emergency braking mode for a few seconds to assure safety. By comparing the results in Figs. \ref{DHSA_Comparision_T_0.7} and \ref{DHSA_GP_T_0.7_loss_0.75}, it is perceived that the performance of the DH-DHSMPC with low-rate communication is very similar to the performance of the DH-MPC with frequent and perfect communication (second case study). On the other hand, as shown in Fig. \ref{DHSA_accident}, the DHMPC is not able to preserve the safety of the platoon and during the sudden deceleration, the second follower vehicle is not able to react properly and hence crashes into its front vehicle.

\section{Conclusion}

\noindent In this paper, a discrete hybrid stochastic MPC design method was proposed for CACC applications by leveraging model-based communication. The proposed method aims at reducing the reliance of the vehicles on communication, i.e., operating safely during communication failure. It was assumed that vehicles share their future acceleration profile as well as an updated model for their speed profile (using Gaussian process) at each successful communication event. For safety purposes, vehicles may operate in either free following mode or emergency braking mode. The vehicle operating mode was chosen based on the predictive speed profile of the preceding vehicle (either received through communication or generated using the last GP model available). The performance of the proposed controller 
was evaluated through simulation studies, which validated the efficacy of the proposed method even with a low-rate intermittent communication.

\section{Acknowledgement}
This research was supported by the National Science Foundation under grants CNS-1931981 and CNS-1932037.

\balance
\bibliography{refs.bib}{}
\bibliographystyle{unsrt}

\end{document}